\documentclass[sigconf]{acmart}
\usepackage{multirow}

\AtBeginDocument{%
  }

\copyrightyear{2025}
\acmYear{2025}
\setcopyright{cc}
\setcctype{by-nc-sa}
\acmConference[MM '25]{Proceedings of the 33rd ACM International Conference on Multimedia}{October 27--31, 2025}{Dublin, Ireland}
\acmBooktitle{Proceedings of the 33rd ACM International Conference on Multimedia (MM '25), October 27--31, 2025, Dublin, Ireland}\acmDOI{10.1145/3746027.3754851}
\acmISBN{979-8-4007-2035-2/2025/10}

\acmSubmissionID{788}

\newcommand{\ie}{{\textit{i.e., }}}
\newcommand{\eg}{{\textit{e.g., }}}
\newcommand{\etal}{\textit{et al.}}

\begin{document}

\title{Relightable and Dynamic Gaussian Avatar Reconstruction \texorpdfstring{\\}{ }
from Monocular Video}

%
\author{Seonghwa Choi}
\affiliation{%
  \institution{Yonsei University}
  \city{Seoul}
 \country{South Korea}
}\email{csh0772@yonsei.ac.kr}

\author{Moonkyeong Choi}
\affiliation{%
  \institution{Yonsei University}
  \city{Seoul}
 \country{South Korea}
}\email{bryan1302@yonsei.ac.kr}

\author{Mingyu Jang}
\affiliation{%
  \institution{Yonsei University}
  \city{Seoul}
 \country{South Korea}
}\email{jmg1002@yonsei.ac.kr}

\author{Jaekyung Kim}
\affiliation{%
  \institution{Yonsei University}
  \city{Seoul}
 \country{South Korea}
}\email{jkkproject@yonsei.ac.kr}

\author{Jianfei Cai}
\affiliation{%
  \institution{Monash University}
  \city{Melbourne}
 \country{Australia}
}\email{Jianfei.Cai@monash.edu}

\author{Wen-Huang Cheng}
\affiliation{%
  \institution{National Taiwan University}
  \city{Taipei}
 \country{Taiwan}
}\email{wenhuang@csie.ntu.edu.tw}

\author{Sanghoon Lee}
\authornote{Corresponding author}
\affiliation{%
  \institution{Yonsei University}
  \city{Seoul}
 \country{South Korea}
}\email{slee@yonsei.ac.kr}

\renewcommand{\shortauthors}{Seonghwa Choi et al.}

\begin{abstract}
Modeling relightable and animatable human avatars from monocular video is a long-standing and challenging task. Recently, Neural Radiance Field (NeRF) and 3D Gaussian Splatting (3DGS) methods have been employed to reconstruct the avatars. However, they often produce unsatisfactory photo-realistic results because of insufficient geometrical details related to body motion, such as clothing wrinkles. In this paper, we propose a 3DGS-based human avatar modeling framework, termed as Relightable and Dynamic Gaussian Avatar (RnD-Avatar), that presents accurate pose-variant deformation for high-fidelity geometrical details. To achieve this, we introduce dynamic skinning weights that define the human avatar's articulation based on pose while also learning additional deformations induced by body motion. We also introduce a novel regularization to capture fine geometric details under sparse visual cues. Furthermore, we present a new multi-view dataset with varied lighting conditions to evaluate relight. Our framework enables realistic rendering of novel poses and views while supporting photo-realistic lighting effects under arbitrary lighting conditions. Our method achieves state-of-the-art performance in novel view synthesis, novel pose rendering, and relighting.
\end{abstract}

\begin{teaserfigure}
  \includegraphics[width=\textwidth]{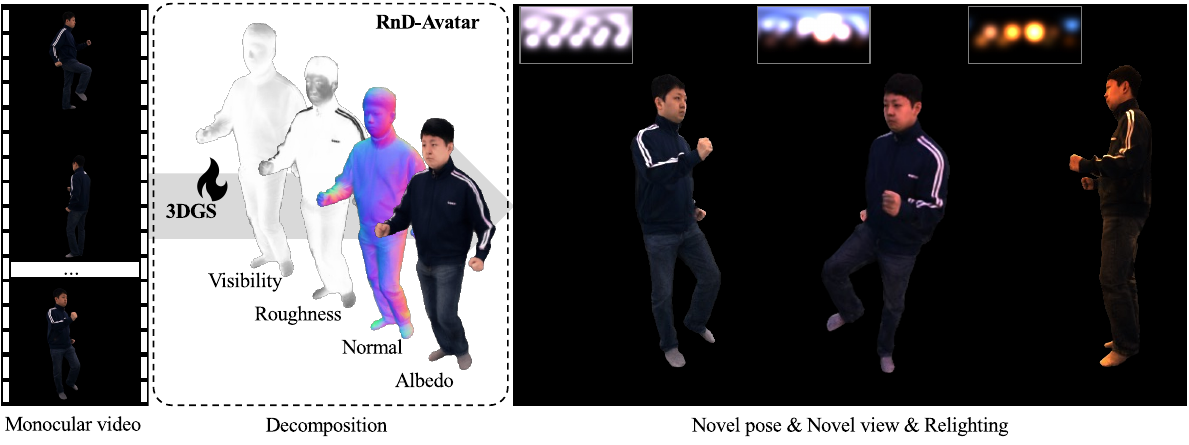}
  \caption{We present a method to reconstruct human avatar from monocular video. Our method decomposes the geometry and appearance attributes of the avatar and can render novel pose/view images under arbitrary lighting conditions.}
  \Description{teasure}
  \label{fig:teaser}
\end{teaserfigure}

\maketitle
\section{Introduction}
Modeling a human avatar from monocular video has attracted significant attention due to its potential for many multimedia applications, including Metaverse, VR/AR, gaming, movies, and virtual try-ons.
Achieving a photo-realistic avatar requires detailed geometry and appearance modeling to render realistic lighting effects under various environmental conditions.
Traditional approaches usually employ specialized equipment, for capturing avatar geometry and appearance, such as well-structured camera systems~\cite{tong2012scanning,bogo2015detailed,collet2015high,habermann2019livecap,xiang2021modeling,xiang2022dressing,xiang2023drivable} or light stages~\cite{guo2019relightables,zhang2021neural,bi2021deep,yang2023towards,sarkar2023litnerf}, which require either professional skills and are labor-intensive work. To alleviate this, in this paper, we focus on modeling human avatars that are both relightable and animatable, using only monocular video input.

Recent works have attempted to represent human avatars by leveraging Neural Radiance Fields (NeRF)~\cite{NeRF} or 3D Gaussian Splatting (3DGS)~\cite{3dgs}, from monocular or multiview videos.
NeRF-based approaches~\cite{peng2021neural, peng2021animatable, liu2021neural,  zhi2022dual, weng2022humannerf, wang2022arah, jiang2022selfrecon,geng2023learning, jiang2023instantavatar, huang2024efficient,lombardi2021mixture, remelli2022drivable, chen2023primdiffusion, chen2022relighting4d,luvizon2023relightable, sun2023neural, lin2024relightable,wang2024intrinsicavatar, xu2024relightable} model the human avatar as implicit representations.
Specifically, these methods infer the geometry and appearance of the avatar by learning inverse mapping correspondences between the canonical and observation spaces using Linear Blend Skinning (LBS). However, this approach often leads to suboptimal rendering results due to ambiguous correspondences between the two spaces.
Moreover, it demands extensive computational resources, leading to slow rendering performance.
In contrast, 3DGS-based~\cite{qian20243dgs,pang2024ash,hu2024gauhuman,shao2024degas,liu2024gea,wen2024gomavatar,kocabas2024hugs,shao2024splattingavatar,jiang2024uv,saito2024relightable,li2024animatable,paudel2024ihuman,moon2024expressive,hu2024expressive} explicitly model avatars with the set of 3D Gaussians.
3DGS-based methods typically leverage a forward LBS to articulate the avatar, following the mechanism of traditional mesh deformation.
While 3DGS-based methods enable accurate detail rendering with lightweight rendering process compared to NeRF-based methods, there are two main limitations to represent fine-detailed geometry of the human avatar: (1) the skinning weight lacks considering complex deformation caused by body motion, such as local clothing deformations, and (2) modeling human avatars from monocular video remains challenging due to limited visual cues that leads to suboptimal optimization of depth-related geometry, such as normal estimation.

\begin{figure}
	\centering
	\includegraphics*[width=0.45\textwidth]{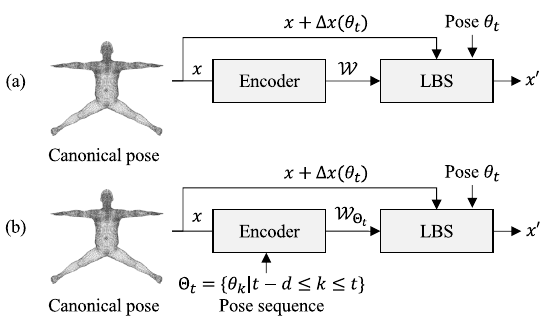}
	\caption{Conceptual comparison between (a) existing 3DGS-based avatar modeling and (b) our approach, where $x$ represents the position of Gaussian, $\mathcal{W}$ denotes skinning weights, $\Theta_{t}$ is a pose sequence at frame $t$, and $\Delta x(\theta_{t})$ represents pose-dependent offset from $\theta_t$. Unlike existing methods, we articulate the human avatar by computing the skinning weights $\mathcal{W}_{\Theta_{t}}$ conditioned on $\Theta_{t}$. For brevity, other attributes of 3DGS are omitted.}
	\label{Fig:approach}
\end{figure}

To address these limitations, we propose a Relightable and Dynamic Gaussian Avatar (RnD-Avatar) that models a human avatar with fine-detailed geometry from a monocular video, leading high quality rendering results with realistic lighting effects.
For modeling a fine-grained avatar, we introduce dynamic skinning weights that enable pose-variant deformation, which is adaptively computed based on motion-dependent conditions, such as body movement.
Fig.~\ref{Fig:approach} illustrates the architectural difference between existing 3DGS-based methods and our approach. As shown in the figure, unlike previous works, we obtain skinning weights $\mathcal{W}_{\Theta_{t}}$ conditioned on both the position of the 3D Gaussian $x$ and pose $\Theta_{t}$, and then articulate the avatar through LBS.
Finally, we optimize RnD-Avatar through a Physically-Based Rendering (PBR) process to optimize geometric attributes (\ie position and normal) and appearance attributes (\ie albedo, roughness, and visibility).
Additionally, we introduce a novel regularization term that facilitates geometry learning from limited visual cues, thereby enhancing depth-related structure for more accurate normal estimation.
This enables realistic lighting effects on the avatar under diverse environmental lighting conditions.

Current the modeling relightable human avatar approaches usually rely only on qualitative performance due to the absence of available datasets that enable quantitative evaluation of relighting performance. To address this gap, we have constructed a dedicated database for relightable human avatar modeling. Unlike existing datasets, our database provides multi-view sequences under various color lighting conditions. Based on our proposed dataset, we demonstrate that our method achieves state-of-the-art performance in novel pose and view synthesis as well as in relighting.
\begin{figure*}[ht!]
	\centering
	\includegraphics*[width=0.95\textwidth]{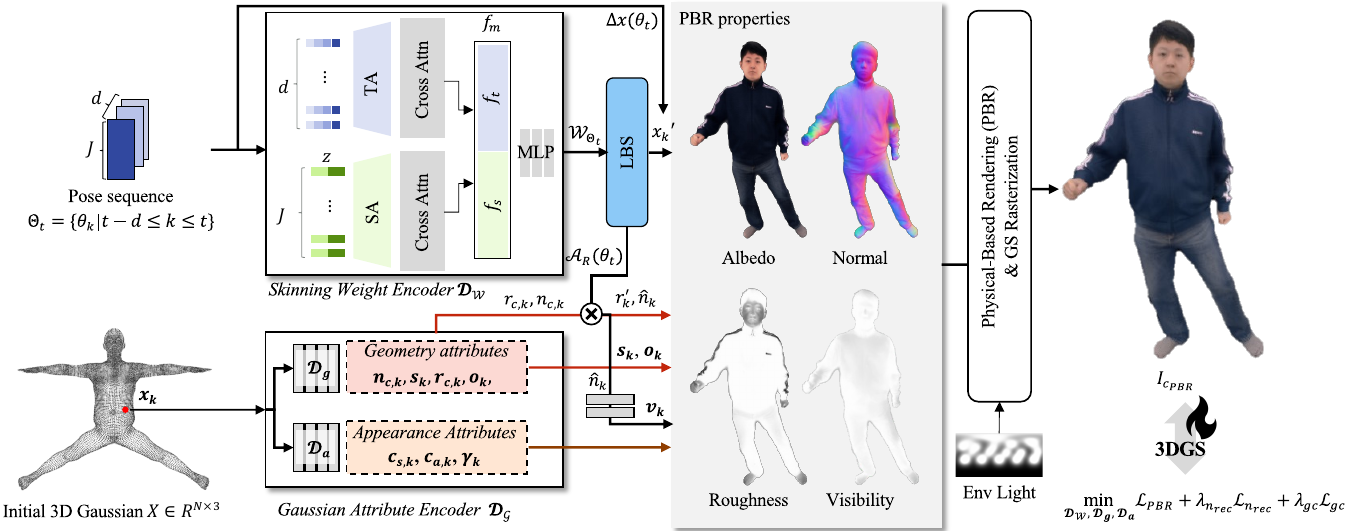}
	\caption{The overall architecture of the proposed method. We initialize the position $x$ of 3D Gaussian using SMPL vertices. Our method produces the dynamic skinning weight $W_{\Theta_{t}}$, which deforms the 3D Gaussians via Linear Blend Skinning (LBS) transformations. To enable relightability, the method learns both geometry and appearance attributes. Finally, given an environmental light, our method renders photorealistic images through a Physically-Based Rendering (PBR) process.}
	\label{Fig:framework}
\end{figure*}
In summary, our contributions in this work include: 
\begin{itemize}
    \item We propose a 3DGS-based human avatar modeling framework, termed Relightable and Dynamic Gaussian Avatar (RnD-Avatar), which reconstructs the animatable and relightable human avatar from a monocular video.
    \item We introduce the dynamic skinning weight to model pose-variant deformations conditioned on body motion. Furthermore, we propose a regularization term to enhance the geometric consistency.
    \item We construct a database for both modeling relightable human avatars and enabling qualitative and quantitative comparisons. The experiments demonstrate the state-of-the-art performance in various tasks of our method, including novel view, novel pose, and relighting.
\end{itemize}


\section{Related work}
\noindent \textbf{Modeling Animatable Human Avatar.}
To model human avatars, some previous methods~\cite{tong2012scanning,collet2015high,bogo2015detailed,habermann2019livecap,xiang2021modeling,xiang2022dressing,xiang2023drivable} have utilized well-structured camera systems that obtain of high-fidelity details of human avatars.
However, these setups are often impractical for real-world applications as they require both professional skills and specialized environments.
Therefore, numerous works~\cite{peng2021neural, peng2021animatable, liu2021neural, zhi2022dual, weng2022humannerf, wang2022arah, jiang2022selfrecon} have explored Neural Radiance Fields (NeRF)~\cite{NeRF}, which models human avatars as an implicit representation.
These approaches typically employ Linear Blend Skinning (LBS)~\cite{smpl} to articulate avatars in implicit neural representations and define an inverse LBS to extract color and density. 
Although they can produce visually appealing rendered avatars, they struggle to capture fine-grained details because indirectly modeling (\eg density) through MLP can lead to suboptimal results and also makes slow inference time.
Some methods~\cite{geng2023learning, jiang2023instantavatar, huang2024efficient} have attempted to enhance training efficiency by leveraging multi-hashing encoding, others~\cite{lombardi2021mixture, remelli2022drivable, chen2023primdiffusion} used neural volumetric primitives for faster rendering; however, achieving both high speed and quality remains a challenge.

To alleviate this, 3D Gaussian Splatting (3DGS)~\cite{3dgs} is an alternative way to address the limitation of NeRF-based methods. 3DGS-based methods~\cite{qian20243dgs,pang2024ash,hu2024gauhuman,li2024animatable,shao2024degas,liu2024gea,wen2024gomavatar,kocabas2024hugs,shao2024splattingavatar,jiang2024uv, paudel2024ihuman,moon2024expressive,hu2024expressive} explicitly represent human avatars as a set of 3D Gaussians, achieving high fidelity rendering results with low computational cost.
Specifically, 3DGS-based methods articulate the human avatar from canonical space to posed space using a forward LBS, which leverages either pre-defined or regressed skinning weights under static conditions (\eg the position of 3D Gaussian in the canonical space).
The fixed skinning weights struggle to capture complex pose-variant deformations, such as clothing wrinkles.
To overcome these limitations, we introduce the dynamic skinning weights for complex geometric deformation.

\noindent \textbf{Modeling Relightable Human Avatar.} 
To model relightable human avatars, NeRF-based methods~\cite{chen2022relighting4d,luvizon2023relightable, sun2023neural, lin2024relightable, xiao2024neca} reconstruct accurate pose-dependent geometry alongside disentangled appearance properties in canonical space by modeling pose-dependent deformation.
To improve the relighting performance, some prior works~\cite{sun2023neural, lin2024relightable} aim to capture fine-grained details by learning the relationship between canonical and observation spaces.
On the other hand, some methods~\cite{wang2024intrinsicavatar, xu2024relightable} have attempted to improve rendering quality by introducing ray tracing~\cite{wang2024intrinsicavatar} within the neural representation for secondary shading effects or a hierarchical distance query algorithm~\cite{xu2024relightable} for generalizing inverse LBS.
However, as mentioned above, due to the limitations of NeRFs with inverse LBS mechanisms, recent 3DGS-based methods for relightable human avatars~\cite{saito2024relightable,li2024animatable} have shown promising results. Nevertheless, they typically require multi-view input videos or a pre-refined mesh from the first frame. In contrast, our method relies solely on a monocular video, enabling a more practical setup for real-world scenarios. Moreover, existing datasets for avatar modeling are limited to evaluate relighting performance due to the lack of ground-truth. To address this, we construct a novel dataset containing multi-view sequences captured under various colored lighting conditions.

\section{Proposed Method}
\label{sec:propoed_method}
\subsection{Overview}
In this section, we describe a Relightable and Dynamic Gaussian Avatar (RnD-Avatar) that models high-quality human avatars using 3D Gaussian Splatting (3DGS). Preliminary details on animatable avatar modeling and 3DGS are provided in the supplementary material. Given a predefined mesh, such as SMPL~\cite{smpl}, we first process its vertices as Gaussian positions to obtain Gaussian attributes, which are categorized into geometry and appearance attributes for modeling a relightable avatar. We introduce dynamic skinning weights that learn pose-variant deformations conditioned on body motion, allowing to articulate the avatar with fine-detailed geometry representation. We render the human avatar through a Physically-Based Rendering (PBR) process and optimize RnD-Avatar with a novel regularization that supplements the sparse visual cues from monocular videos. The detail of our proposed method is shown in Fig~\ref{Fig:framework}.

\subsection{Relightable and Dynamic Gaussian Avatar}
\noindent \textbf{Gaussian Attributes Encoder $\mathcal{D}_{\mathcal{G}}$.} We initially set 3D Gaussian attributes based on SMPL vertices $X\in\mathbb{R}^{N_{g}\times3}$, where $N_{g}$ denotes the total number of vertices. The $k$th Gaussian attributes in $X$ is defined by the position of Gaussian $x_{c,k}$, opacity $o_{k}$, rotation quaternion $r_{c,k}$, scaling factor $s_{k}$, normal vector $n_{c,k}$, the Spherical Harmonics (SH) coefficient for RGB color $c_{k,s}$, albedo color $c_{k,a}$, and roughness $\gamma_{k}$.
We categorize these Gaussian attributes into geometric attributes ($o_{k}, r_{k}, s_{k},$ and $ n_{k}$) and appearance attributes ($c_{k,s}, c_{k,a},$ and $\gamma_{k}$).
Specifically, to obtain the attributes, we design $\mathcal{D}{\mathcal{G}}$, which consists of two sub-encoders: $\mathcal{D}_{g}$ and $\mathcal{D}_{a}$. Both encoders are implemented as multi-layer perceptrons (MLPs) that take the Gaussian position $x$ as input and output the corresponding geometric and appearance attributes, respectively.


\noindent \textbf{Dynamic Skinning Weights Encoder $\mathcal{D}_{\mathcal{W}}$.}
An intuitive way to articulate the human avatar is to apply Linear Blend Skinning (LBS) transformation by using skinning weights which are pre-defined or regressed from static conditions, such as the position of Gaussians in canonical space.
However, these static skinning weights struggle to capture pose-related deformations, such as clothing wrinkles.
To address this, we design $\mathcal{D}_{\mathcal{W}}$ to dynamically compute the skinning weights conditioned on the input pose sequence.
We deform the human avatar through blending skinning transformation computed from dynamic skinning weight $\mathcal{W}_{\Theta_{t}}$.
Since $\mathcal{W}_{\Theta_{t}}$ is influenced by the body motion, RnD-Avatar can model pose-variant deformation.

Given a $d$-length pose sequence at frame $t$, $\Theta_{t} \in \mathbb{R}^{d \times J \times 3} = \{\theta_{t-d}, \ldots, \theta_{t}\}$, we encode the input motion to temporal feature $f_{t}$ and spatial feature $f_{s}$ respectively. Subsequently, we encode the position feature $f_{x}$ through MLPs.
To obtain $f_{t}$, we first reshape the input motion as $f_{p} \in \mathbb{R}^{J \times (d \times z)}$, representing the movement of each joint, representing global motion dynamics. This reshaped motion is then fed into a temporal attention layer.
Next, we apply a cross-attention mechanism, where the key and value are derived from the output of the temporal attention layer, while $f_{x}$ serves as the query. Similarly, to obtain $f_{s}$, we embed the spatial attention output along with $f_{x}$. Specifically, we compute the joint difference between $\theta_{t}$ and $\theta_{t-1}$, which is then fed into a spatial attention layer to capture local motion dynamics.
Finally, we concatenate the two features and feed them into several layers of MLPs to compute the dynamic skinning weight $\mathcal{W}_{\Theta_{t}}$.
The details of $\mathcal{D}_{\mathcal{W}}$ are shown in Fig~\ref{Fig:framework}. 
Based on $\mathcal{D}_{\mathcal{W}}$, we obtain the pose-driven transformation matrices $\mathcal{A}$ which comprises rotation $\mathcal{A}_{R}(\Theta_{t})$ and translation $\mathcal{A}_{T}(\Theta_{t})$, given $\Theta_{t}$ as similar to Eq.~\ref{deformation}: $\mathcal{A}_{T}(\Theta_{t})=\big[\mathcal{A}_{R}(\Theta_{t});\mathcal{A}_{T}(\Theta_{t})\big]$. The position $x$, rotation $r$, and normal $n$ of each 3D Gaussian are dynamically transformed as follows:
\begin{equation}
    \begin{aligned}
    x' &= \mathcal{A}_{R}(\Theta_{t}) \cdot (x_{c}+\Delta x(\theta_{t})) + \mathcal{A}_{T}(\Theta_{t}), \\
    r'&=\mathcal{A}_{R}(\Theta_{t}) \cdot (r_{c}+\Delta r(\theta_{t})),\\
    \hat{n}&=\mathcal{A}_{R}(\Theta_{t}) \cdot n_{c},
    \end{aligned}
\end{equation}
We compute two offset $\Delta x(\theta_{t})$ and $\Delta r(\theta_{t})$ to account for non-rigid deformations through MLPs conditioned on $x$ and $\theta_{t}$. In addition, we also transform the normal vector $n_{c}$ in the cannonical space into the observation space.

\begin{figure}
	\centering
	\includegraphics*[width=0.45\textwidth]{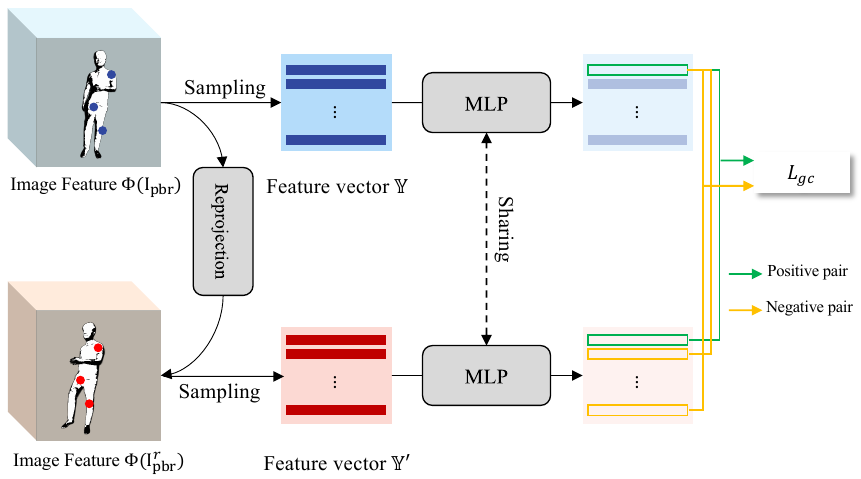}
	\caption{Conceptual visualization of geometric consistency loss. Given two feature maps of rendered images, feature vectors are sampled within the intersection area. Subsequently, our method aims to increase the similarity between positive pairs while reducing the similarity between negative pairs.}
	\label{Fig:gc_loss}
\end{figure}


\noindent \textbf{Physically-Based Rendering (PBR)}. 
We utilize the rendering equation \cite{kajiya1986rendering} to simulate the human avatar under a lighting condition.
The Gaussian attributes are applied to the rendering equation (Eq.~\ref{pbr}) as follows:
\begin{equation}
    \label{pbr}
    L_{o}(x, \omega_{o})=
    \sum_{\omega_{i}}L_{i}(x,\omega)f(\omega_{i}, \omega_{o})(\omega_{i} \cdot \hat{n}(x))\Delta \omega_{i},
\end{equation}

\noindent where $L_{i}(x,\omega_{i})$ and $L_{o}(x,\omega_{o})$ are the incident and outgoing radiance at a position $x$ along direction $\omega_{i}$ and $\omega_{o}$, respectively. $L_{i}(x,\omega_{i})$ is computed by the visibility $v$ and a global light $L(\omega_{i})$ at each Gaussian: $L_{i}(x,\omega_{i})=v(x,\omega_{i})L(\omega_{i})$. We parameterize the visibility $v(x,\omega_{i})$ by 3-degree SH coefficients to present a mono channel. 
Specifically, we implement view ($\omega^{c}_i=c-x$)-dependent visibility, modeled as a SH function: $v(x, \omega^{c}_i)=\sum_j v_j Y_j(\omega^{c}_i)$. It is computed in canonical space via a lightweight MLP using $x_c$ and $\hat{n}_o$ to handle pose variation.
$L(\omega_{i})$ is an environment light as a learnable light probe in latitude-longitude format $\in \mathbb{R}^{32\times64\times1}$.
We employ the Disney Bidirectional Reflectance Distribution Function (BRDF)~\cite{burley2012physically} $f(\cdot)$, influenced by the albedo $c_{a}$, normal $n$, roughness $\gamma$, and metallic. We manually set the metallic value to zero to simplify the modeling of geometry and appearance attributes.
Thus, the PBR can be defined as follows:
\begin{equation}
    \label{brdf}
    f(\omega_{i}, \omega_{o}) = \frac{c_{a}}{\pi} + \frac{D \cdot F \cdot G}{4(n \cdot \omega_{i})(n \cdot \omega_{o})}.
\end{equation}
\noindent where microfacet distribution function $D$, Fresnel reflection $F$, and geometric shadowing factor $G$. $D$ and $G$ are influenced by the roughness $\gamma$. We shade the avatar's color through Eq.~\ref{pbr}, and render the posed human avatar with geometry and appearance attributes through 3DGS rasterization process. 

\subsection{Training}
We train our proposed method in two stages to effectively learn the geometric and appearance properties.

\textbf{First Stage.} We train the geometry attributes (\ie $o,s,r$ and $n$) by optimizing $\mathcal{D}_{\mathcal{W}}$ and $\mathcal{D}_{g}$.
We set the objective function $\mathcal{L}_{stage1}$ consists of the reconstruction loss $\mathcal{L}_{rec}$, and normal reconstruction loss $\mathcal{L}_{n_{rec}}$ as follows:
\begin{equation}
    \label{loss_stage1}
    \mathcal{L}_{stage1}= \mathcal{L}_{rec} + \lambda_{n_{rec}}\mathcal{L}_{n_{rec}},
\end{equation}
\noindent \textit{Reconstruction loss $\mathcal{L}_{rec}$.} Training only with normal vectors often results in low quality, as the lack of visual information makes the training process highly under-constrained. We set the additional color attributes $c_{s}\in\mathbb{R}^{3}$ to guide training the geometry. We note that $I_{c_{s}}$ is the rendered image from $c_{s}$.
We employ $L1$ loss and LPIPS loss~\cite{zhang2018unreasonable} as:
 \begin{equation}
    \label{loss_recon}
    \mathcal{L}_{rec}= \mathcal{L}_{1}(I^{gt}_{rgb}, I_{c_{s}}) + \lambda_{lpips}\mathcal{L}_{lpips}(I^{gt}_{rgb}, I_{c_{s}}),
\end{equation}

\noindent \textit{Normal reconstruction loss $\mathcal{L}_{n_{rec}}$.}
We utilize an off-the-shelf normal estimation network to guide the deformed normal.
Let $I^{gt}_{n}$ is a predicted normal from the network, and $I_{\hat{n}}$ is rasterized image using $\hat{n}$. We compute L1 loss between $I^{gt}_{n}$ and $I_{\hat{n}}$ them as:
\begin{equation}
    \label{loss_normal}
    \mathcal{L}_{n_{rec}}= \mathcal{L}_{1}(I^{gt}_{n}, I_{\hat{n}}),
\end{equation}

\textbf{Second Stage.} The PBR-related optimizable parameters (\textit{i.e.,} $c_{a}$, $v$, $\gamma$, and L) are jointly trained with the geometry attributes.
We remove $c_{s}$ during this stage, which also eliminates $\mathcal{L}_{rec}$.
Hence, we set the objective function by incorporating the normal loss $\mathcal{L}_{n_{rec}}$, PBR loss $\mathcal{L}_{pbr}$, and geometric consistency loss $\mathcal{L}_{gc}$. 
\begin{equation}
    \label{loss_stage2}
    \mathcal{L}_{stage2}= \mathcal{L}_{pbr} + \lambda_{n_{rec}}\mathcal{L}_{n_{rec}} + \lambda_{gs}\mathcal{L}_{gs},
\end{equation}
\noindent \textit{PBR loss $\mathcal{L}_{pbr}$.}
We minimize the difference between the ground-truth RGB $I^{gt}_{rgb}$ and a rendered image $I_{c_{pbr}}$ as:
\begin{equation}
    \label{loss_pbr}
    \mathcal{L}_{pbr}= \mathcal{L}_{1}(I^{gt}_{rgb}, I_{c_{pbr}}) + \lambda_{lpips}\mathcal{L}_{lpips}(I^{gt}_{rgb}, I_{c_{pbr}}),
\end{equation}

\noindent \textit{Geometric consistency loss $\mathcal{L}_{gc}$.}
Modeling a human avatar from monocular video input often results in suboptimal quality due to depth ambiguity caused by the sparsity of visual information. To address this, we design geometric consistency loss $\mathcal{L}_{gc}$, which maximizes the similarity between the training viewpoint and a randomly generated virtual viewpoint. 

Specifically, we begin by rendering an additional image with a randomly augmented virtual camera, $I^{r}_{c_{s}}$, and extract deep representations of both images using a pre-trained network, $\Phi$, such as VGG-16 \cite{simonyan2014very}. We then randomly sample $N$ feature vectors, $\mathbb{Y}^{l} = \{y_{0},\ldots,y_{N}\}^{i}$ and $\mathbb{Y'}^{l} = \{y'_{0},\ldots,y'_{N}\}^{i}$, from the $i$th layers of features for $I_{c_{pbr}}$ and $I^{r}_{c_{pbr}}$, respectively. These feature vectors are sampled within commonly visible regions, allowing us to compare corresponding points between the two sets. Our network is trained to increase the similarity of matching points (\textit{i.e.,} $(y_{i}, y'_{j})$, where, $i=j$) while decreasing the similarity of contrasting points (\textit{i.e.,} $(y_{i}, y'_{j})$, where, $i\neq j$).
To achieve this, we formulate $\mathcal{L}_{gc}$ using InfoNCE loss~\cite{oord2018representation} as follows:

\begin{equation}
    \label{loss_gc}
    \mathcal{L}_{gc}\! =\! \sum_{i=1}^{S_{l}}\! \sum_{j=1}^{S_{p}} -\!\log\!\! \left(\! \frac{\exp \left( y_j^i \cdot \mathbb{Y'}^i \right)}{\exp\!\left( y_j^i\! \cdot\! \mathbb{Y'}^i \right)\!\! +\! \!\sum_{k=1, k \neq j}^{N} \exp\! \left( y_k^i\! \cdot\! \mathbb{Y'}^i \right)\!\!} \right)\!.
\end{equation}


\noindent where $S_{l}$ is the set of layers selected from $\Phi$, and $S_{p}$ is the index set of generated points.
    
\section{Experiments}
\begin{figure*}[ht!]
	\centering
	\includegraphics*[width=0.95\textwidth]{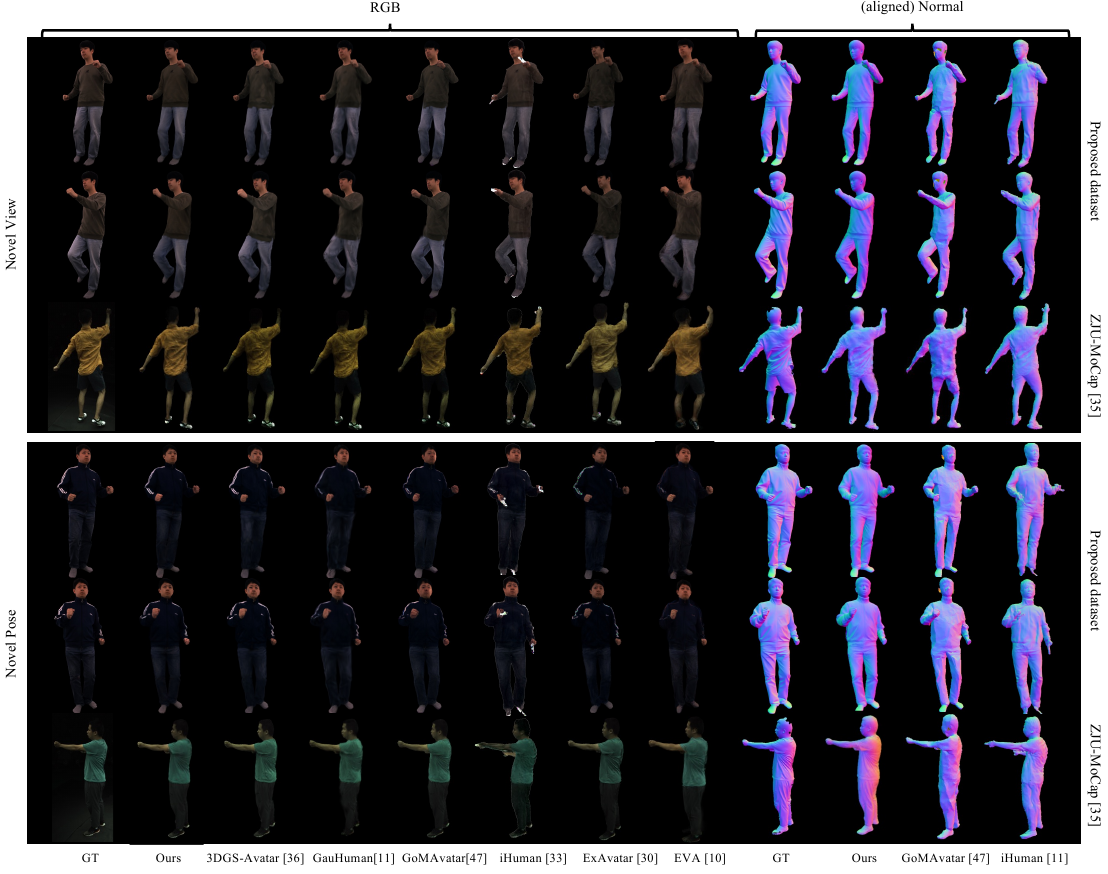}
	\caption{Qualitative results of human avatar reconstruction (novel pose and view rendering under white environmental light).}
	\label{Fig:qa_recon}
\end{figure*}

\begin{figure}[ht!]
	\centering
	\includegraphics*[width=0.45\textwidth]{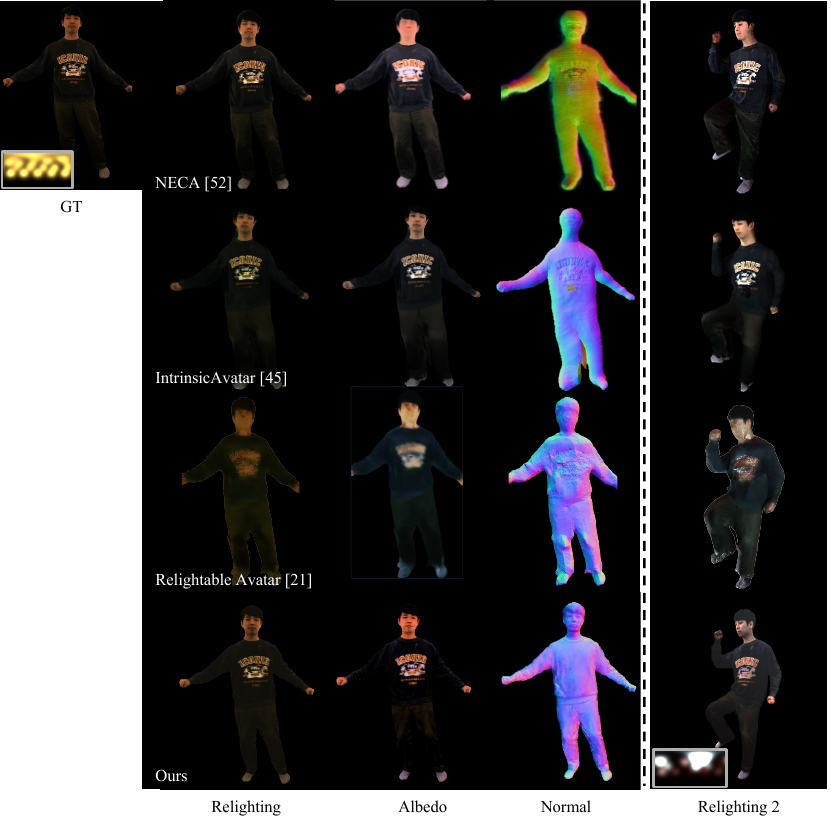}
	\caption{Qualitative results of human avatar relighting.}
	\label{Fig:qa_relight_mdi1}
\end{figure}

\begin{figure}[ht!]
	\centering
	\includegraphics*[width=0.40\textwidth]{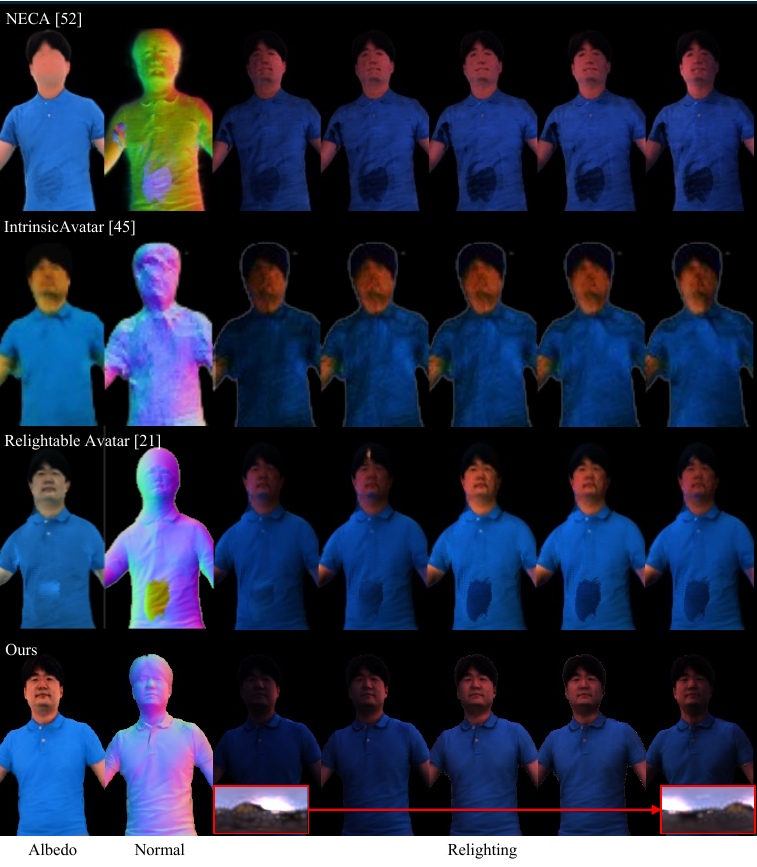}
	\caption{Qualitative results of human avatar relighting.}
	\label{Fig:qa_relight_dynamic1}
\end{figure}




\begin{figure}
	\centering
	\includegraphics*[width=0.43\textwidth]{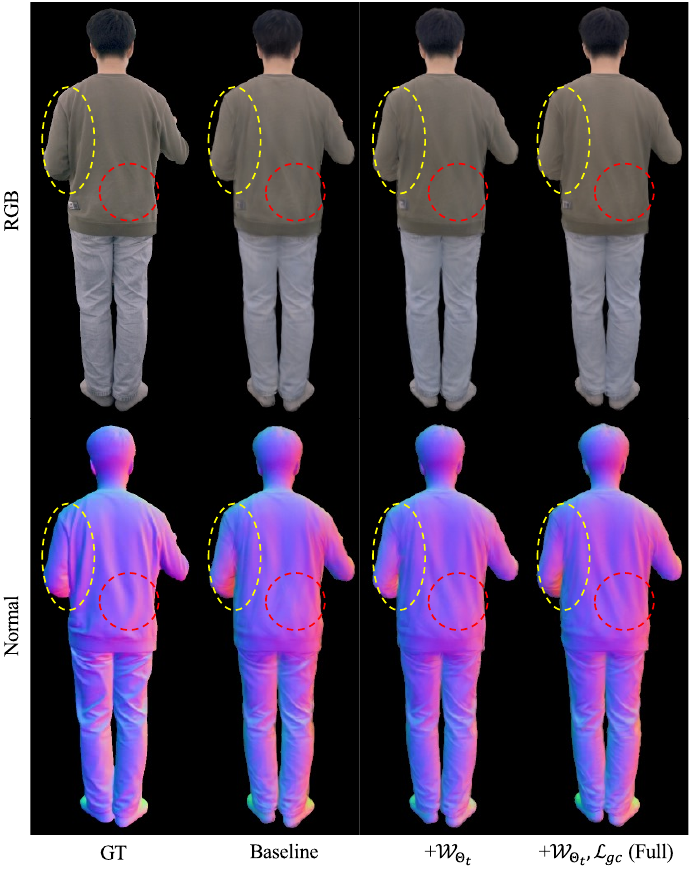}
	\caption{Ablation study of $\mathcal{W}_{\Theta_{t}}$ and $\mathcal{L}_{gc}$.}
	\label{Fig:ablation}
\end{figure}

\begin{figure}
	\centering
	\includegraphics*[width=0.42\textwidth]{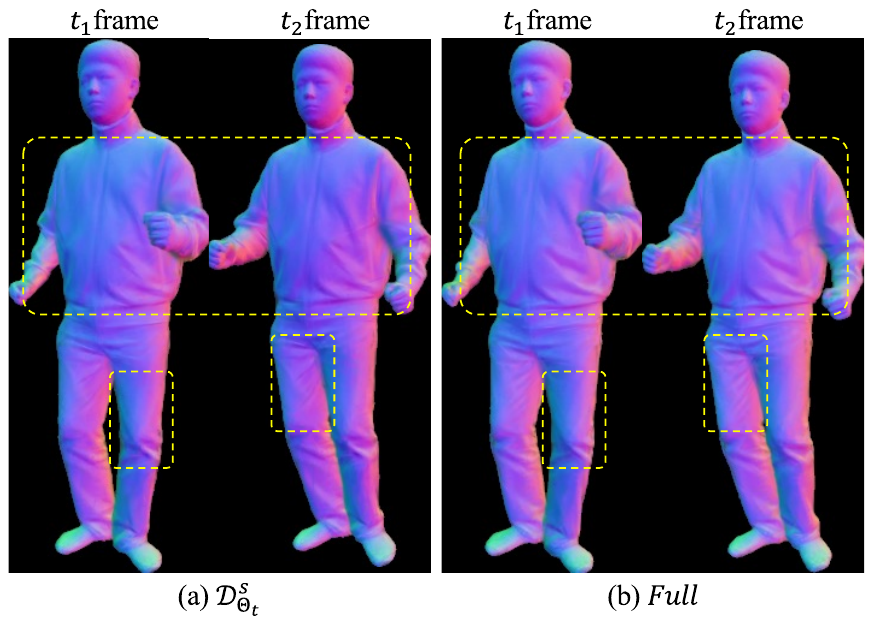}
	\caption{Ablation study of the skinning weight encoder.}
	\label{Fig:ablation_module}
\end{figure}



\begin{table}[]
    \centering
    \caption{Comparison of multi-view human datasets.}
    
    \resizebox{0.45\textwidth}{!}{
    \begin{tabular}{c|ccccc}
    \hline \hline
    Dataset    & \#ID   & \#View &  \#Light Color &\#Frames & Resolution \\ \hline
    Human3.6M \cite{ionescu2013human3}   &  11    &    4        &     1        &  3.6M   &  1000P     \\
    MPI-INF-3DHP \cite{mehta2017monocular}&  8     &    14     &     1        &  1.3M   &  2048P     \\
    ZJU-MoCap \cite{peng2021neural}   &  10    &    24     &     1        &  180K   &  1024P     \\
    THuman 4.0 \cite{zheng2022structured}  &   3    &    24        &     1        &   10K   &  1150P     \\
    RDA \cite{relightneuralactor2024eccv} &   4    &    \textbf{8 - 100}        &     1        &   90K   &  1024P     \\
    Ours        &  \textbf{20}    &    \textbf{30}      &     \textbf{8}        &  \textbf{11.5M}  &  \textbf{4096P}     \\ \hline \hline
    
    \end{tabular}
    }
    \label{table:dataset}
    \end{table}

\subsection{Multi-view Multi-illuminated Dataset}
Existing human performance datasets \cite{ionescu2013human3, mehta2017monocular, peng2021neural, zheng2022structured, relightneuralactor2024eccv} for modeling human avatars typically capture subjects under white lighting, making it challenging to evaluate relighting accuracy by comparing it with real ground truth. While synthetic datasets offer an alternative, they often suffer from geometric and appearance artifacts that affect relighting evaluation. To address this limitation, Luvizon~\etal~\cite{relightneuralactor2024eccv} introduced a dataset that captures the human under six indoor and one outdoor lighting conditions. The dataset primarily varies lighting direction, but it still retains a restricted color range, making it challenging to comprehensively evaluate relighting performance.
To bridge this gap, we present a novel dataset comprising eight subjects performing various actions under a diverse range of lighting colors. The detail of proposed dataset is described in the supplementary material.

\subsection{Qualitative Results}
\noindent \textbf{Comparison Methods.}
We compare with state-of-the-art methods for both human avatar reconstruction (3DGS-Avatar~\cite{qian20243dgs}, Gauhuman~\cite{hu2024gauhuman}, GomAvatar~\cite{wen2024gomavatar}, iHuman~\cite{paudel2024ihuman}, ExAvatar~\cite{moon2024expressive}, and EVA~\cite{hu2024expressive}) and relightable avatar modeling (RelightableAvatar~\cite{lin2024relightable}, NECA~\cite{xiao2024neca}, and IntrinsicAvatar~\cite{wang2024intrinsicavatar}).
The existing methods take a monocular video and a 3D pose as input and produce a human avatar capable of novel pose and novel view rendering.
We train these methods according to their original training procedures, adapted to our training dataset setup.
The implementation detail is described in the supplementary material.

\noindent \textbf{Human Avatar Reconstruction.}
We conduct a qualitative comparison to assess the reconstruction performance against the human avatar reconstruction~\cite{qian20243dgs,hu2024gauhuman,wen2024gomavatar,paudel2024ihuman,moon2024expressive,hu2024expressive}.
We trained both the benchmark methods and our approach on our proposed dataset as well as the ZJU-Mocap dataset~\cite{peng2021neural}.
The results are shown in Fig.~\ref{Fig:qa_recon}.
We render the human avatar in both novel poses and views under white environmental lighting.
Specifically, GoMAvatar~\cite{wen2024gomavatar} and iHuman~\cite{paudel2024ihuman} can render the normal of avatar, so we also compare the quality of the rendered normal maps. As shown in the figure, they struggle to preserve the geometric coherency when the avatar is animated, and show insufficient local details and visually unpleasing deformations, such as clothing wrinkles. In contrast, our method demonstrates superior detailed results with consistency.

\noindent \textbf{Relightable Avatar Modeling.}
For qualitative evaluation of relighting, we compared our method with relightable avatar modeling methods~\cite{lin2024relightable, xiao2024neca, wang2024intrinsicavatar}.
Both our method and the relighting methods were trained on our proposed dataset as well as the ZJU-Mocap dataset~\cite{peng2021neural}.
Fig.~\ref{Fig:qa_relight_mdi1} shows estimated albedo, normal and relighting results to compare the relighting performance.
Additionally, \textit{Relighting 1} represents the rendered result with a novel pose, while \textit{Relighting 2} depicts the rendered avatar under arbitrary lighting with a novel view.
Our proposed dataset allows us to directly compare the result (\textit{Relighting 1}) with the ground truth.
Additional results on the ZJU-Mocap dataset are provided in the supplementary material.
We observed that NECA~\cite{xiao2024neca} and RelightingAvatar~\cite{lin2024relightable} produce visually appealing results; however, the geometrical details appear inaccurate. Furthermore, as shown in the figure, IntrinsicAvatar~\cite{wang2024intrinsicavatar} struggles to reconstruct detailed avatars due to the wide range of body pose variations in our proposed dataset. This limitation arises from inferring geometry based on density by an implicit manner. 
Furthermore, we present an additional comparison of relighting performance by rendering human avatars under dynamic lighting in Fig.~\ref{Fig:qa_relight_dynamic1}. 
Thanks to modeling the accurate normal details, RnD-Avatar demonstrates smooth variations in lighting effects compared to existing methods. 
\begin{table*}[t]
    \centering
    \caption{Quantitative results of human avatar reconstruction on our database. ``$\uparrow$" indicates higher is better. ``$\downarrow$" indicates the opposition.}
    \resizebox{0.85\textwidth}{!}{
    \begin{tabular}{c|ccccc|ccccc|c}
    \hline\hline
    \multirow{2}{*}{Method} & \multicolumn{5}{c|}{Novel view} & \multicolumn{5}{c|}{Novel pose} & \multirow{2}{*}{TR} \\
                             & PSNR$\uparrow$   & SSIM$\uparrow$   & LPIPS$\downarrow$ & PSNR$_{n}$$\uparrow$   & SSIM$_{n}$$\uparrow$    & PSNR$\uparrow$   & SSIM$\uparrow$   & LPIPS$\downarrow$ & PSNR$_{n}$$\uparrow$   & SSIM$_{n}$$\uparrow$    \\ \hline
                             3DGS-Avatar\cite{qian20243dgs}                   &30.85&0.9476&0.0266&-&-               &29.12&0.9376&0.0316&-&-& 2h\\
                             GauHuman\cite{hu2024gauhuman}                    &29.34&0.9386&0.0239&-&-               &27.26&0.9021&0.0389&-&-& 1h\\
                             GoMAvatar\cite{wen2024gomavatar}                 &31.13&0.9490&0.0191&18.35&0.7614      &29.73&0.9328&0.0291&17.15&0.7172& 12h\\
                             iHuman\cite{paudel2024ihuman}                    &26.87&0.8848&0.0342&16.42&0.6248      &24.84&0.8691&0.0453&13.15&0.5894& 2h\\
                             ExAvatar\cite{moon2024expressive}                &30.28&0.9457&0.0171&-&-               &30.17&0.9344&0.0301&-&-& 2h\\
                             EVA\cite{hu2024expressive}                       &30.42&0.9346&0.0267&-&-               &28.61&0.9217&0.0332&-&-& 6h\\
                             \hline
                             
                             Ours                                             &\textbf{31.92}&\textbf{0.9621}&\textbf{0.0150}&\textbf{26.94}&\textbf{0.9509} &\textbf{30.19}&\textbf{0.9427}&\textbf{0.0270}&\textbf{26.48}&\textbf{0.9487}& 6h\\ \hline \hline
    \end{tabular}
    }
    \label{table:quant_recon}
\end{table*}

\begin{table}[]
    \centering
    \caption{Quantitative results of relighting human avatar under color environmental light on our database. ``$\uparrow$" indicates higher is better. ``$\downarrow$" indicates the opposition.}
    \resizebox{0.45\textwidth}{!}{
    \begin{tabular}{c|ccccc|c}
    \hline\hline
    \multirow{2}{*}{Method} & \multicolumn{5}{c|}{Novel view} & \multirow{2}{*}{TR} \\
    & PSNR$\uparrow$   & SSIM$\uparrow$   & LPIPS$\downarrow$ & PSNR$_{n}$$\uparrow$   & SSIM$_{n}$$\uparrow$    \\ \hline
    NECA~\cite{xiao2024neca}     &30.52&0.9198&0.0416&17.87&0.6492& 4h\\
    RelightableAvatar~\cite{lin2024relightable}   &28.26&0.8956&0.0589&17.95&0.6541& 8h\\
    IntrinsicAvatar~\cite{wang2024intrinsicavatar}  &28.26&0.8956&0.0589&15.14&0.5475& 12h\\ \hline
    Ours    &\textbf{30.78}&\textbf{0.9231}&\textbf{0.0363}&\textbf{26.94}&\textbf{0.9509}  & 6h\\ \hline \hline
    \end{tabular}
    }
    \label{table:quant_relight}
\end{table}


\subsection{Quantitative Results}
For a quantitative evaluation, we compared PSNR, SSIM, and LPIPS to evaluate the fidelity of rendered results across novel pose, novel view, and relighting tasks.
Furthermore, to evaluate geometry quality, we compute the fidelity of rendered normal results using PSNR and SSIM, denoted as PSNR$_{n}$ and SSIM$_{n}$, respectively. Additionally, we report the training time (TR) required to optimize a human avatar.
Tab.~\ref{table:quant_recon} reports the quantitative comparison of human avatar reconstruction, while Tab.~\ref{table:quant_relight} presents the performance of relighting human avatar.
As shown, our method achieves higher fidelity in both reconstruction, relighting, and geometrical detail (normal quality) compared to existing approaches.
These results indicate that our proposed method performs effective human avatar articulation and achieves photorealistic rendering quality.

\subsection{Ablation Study}
We additionally conducted ablation experiments to verify the contributions of our proposed framework. To this end, we established a \textit{Baseline} architecture consisting of an MLP and static skinning weights for articulation.

\noindent \textbf{Effectiveness of Dynamic skinning weight.}
We conduct an ablation study to show the effectiveness of the proposed dynamic skinning weights. Specifically, we compared the performance between \textit{Baseline} and \textit{Baseline}+$\mathcal{W}_{\Theta_{t}}$.
The qualitative and quantitative results are shown in Fig.~\ref{Fig:ablation} and Tab.~\ref{table:ablation}.
As shown in the results, we can evidence that $\mathcal{W}_{\Theta_{t}}$ enhance the geometry details.

\noindent \textbf{Effectiveness of Geometric Consistency Loss.}
An ablation study was conducted to verify the effectiveness of $\mathcal{L}_{gc}$. 
Qualitative comparisons are shown in Fig.~\ref{Fig:ablation}, where we focus on relighting results to illustrate the influence of geometry quality.
Specifically, we observe that the human avatar without $\mathcal{L}_{gc}$ results in an inaccurate surface representation of the human avatar.
Furthermore, as shown in Tab.~\ref{table:ablation}, using $\mathcal{L}_{gc}$ results in higher fidelity outputs compared to when it is not applied. This suggests that our regularization significantly improves the geometry.

\noindent \textbf{Skinning Weight Encoder Variants.}
To compute $\mathcal{W}_{\Theta_{t}}$, we leverage the pose sequence to capture both global and local motion dynamics.
We conducted an ablation study to validate the encoder’s capability in capturing both global and local motion dynamics ($f_{t}$ and $f_{s}$) for accurate skinning weight estimation.
We design an encoder $\mathcal{D}^{S}_{\Theta_{t}}$ that generates the skinning weights solely from $f_{s}$.
Fig.~\ref{Fig:ablation_module} shows the estimated normals at two consecutive frames ($t_{1}$ and $t_{2}$).
As shown in the figure, although the body rotation is small, Fig.\ref{Fig:ablation_module} (a) exhibits a noticeable change in the overall orientation of the normal vectors, particularly in the chest region, while Fig.\ref{Fig:ablation_module} (b) can address this limitation while achieves higher normal fidelity as reported in Table.~\ref{table:ablation_module}. This suggests that computing skinning weights requires consideration of both global and local motion dynamics.

\noindent \textbf{Influence of Pose Sequence $d$.}
We further explored the influence of the pose sequence $d$ used to compute $f_{m}$.
We report the quantitative performance of novel-view synthesis while varying the sequence length.
Additionally, we compute the floating point operations (FLOPs) of the pose encoder based on the sequence length. As shown, the performance with a sequence length of 10 and 20 is comparable; however, a length of 20 yields slightly more optimal results. Nevertheless, the computational cost increases significantly with longer sequences. Therefore, we set $d = 10$ in our experiments to balance performance and efficiency.

\begin{table}[]
\centering
\caption{Quantitative result of ablation study. ``$\uparrow$" indicates higher is better. ``$\downarrow$" indicates the opposition.}

\resizebox{0.4\textwidth}{!}{
\begin{tabular}{l|ccccc}
\hline \hline
\multirow{2}{*}{method} & \multicolumn{5}{c}{Novel view} \\
                        & PSNR$\uparrow$   & SSIM$\uparrow$   & LPIPS$\downarrow$   & PSNR$_{n}$$\uparrow$   & SSIM$_{n}$$\uparrow$ \\ \hline
\textit{baseline}           &29.73&0.9311&0.0251&24.83&0.9413 \\
$+\mathcal{W}_{\Theta_{t}}$ &30.96&0.9456&0.0189&25.51&0.9456 \\
$+\mathcal{L}_{gc}$ (\textit{Full}) &\textbf{31.92}&\textbf{0.9621}&\textbf{0.0150}&\textbf{26.48}&\textbf{0.9487} \\ \hline \hline
\end{tabular}
}
\label{table:ablation}
\end{table}

\begin{table}[]
    \centering
    \caption{Quantitative result of ablation study. ``$\uparrow$" indicates higher is better. ``$\downarrow$" indicates the opposition.}
    
    \resizebox{0.2\textwidth}{!}{
    \begin{tabular}{l|cc}
    \hline \hline
    \multirow{2}{*}{method} & \multicolumn{2}{c}{Novel view} \\
                            & PSNR$_{n}$$\uparrow$   & SSIM$_{n}$$\uparrow$ \\ \hline
    $\mathcal{D}^{S}_{\Theta_{t}}$&26.48&0.9487 \\
    \textit{Full} &\textbf{26.94}&\textbf{0.9509} \\ \hline \hline
    \end{tabular}
    }
    \label{table:ablation_module}
\end{table}

\begin{table}[]
\centering
\caption{Quantitative novel-view synthesis results based on varying pose sequence length $d$. ``$\uparrow$" indicates higher is better. ``$\downarrow$" indicates the opposition.}
\resizebox{0.35\textwidth}{!}{
\begin{tabular}{c|ccccc}
\hline \hline

$d$&2&5&10&20&30 \\\hline 
PNSR $\uparrow$&29.32&31.81&\textbf{31.92}&32.02&30.51 \\ 
SSIM $\uparrow$&0.9387&0.948&\textbf{0.962}&0.965&0.951 \\
FLOPs(G)$\downarrow$&0.016&0.035&\textbf{0.054}&0.102&0.149 \\\hline \hline

\end{tabular}
}
\label{table:ablation_d}
\end{table}

\section{Conclusion}
In this paper, we propose RnD-Avatar, a method designed to model detailed human avatars for rendering novel poses/views and enabling relighting under arbitrary environmental light. The core of our approach is to learn the pose-variant deformation for the fine-grained geometric details of human avatar from monocular videos. To achieve this, we introduce a dynamic skinning weight that leverages input body motion (dynamic) to compute pose-variant transformation matrices.
This is used for skeleton-driven deformation and modeling fine-grained geometry for avatar articulation.
Furthermore, to address the sparsity of monocular videos, we introduce a novel regularization that enhances geometric consistency. Additionally, we construct a database that captures human motion videos under diverse lighting conditions. Leveraging this database, our method demonstrates state-of-the-art performance in tasks such as novel view synthesis, novel pose rendering, and relighting.

\begin{acks}
This research was supported by Culture, Sports and Tourism R\&D Program through the Korea Creative Content Agency grant funded by the Ministry of Culture, Sports and Tourism in 2024 (RS-2024-00398413, Contribution Rate: 85\%), and the National Research Foundation of Korea (NRF) grant funded by the Korea government (MSIT) (RS-2025-02216328, Contribution Rate: 15\%).
\end{acks}

\bibliographystyle{ACM-Reference-Format}
\bibliography{sample-base}

\clearpage
\newpage
\appendix

In this supplementary, we provide more detailed descriptions and experimental results of our proposed framework.
\section{Appendix}
\begin{figure*}[ht!]
	\centering
	\includegraphics*[width=\textwidth]{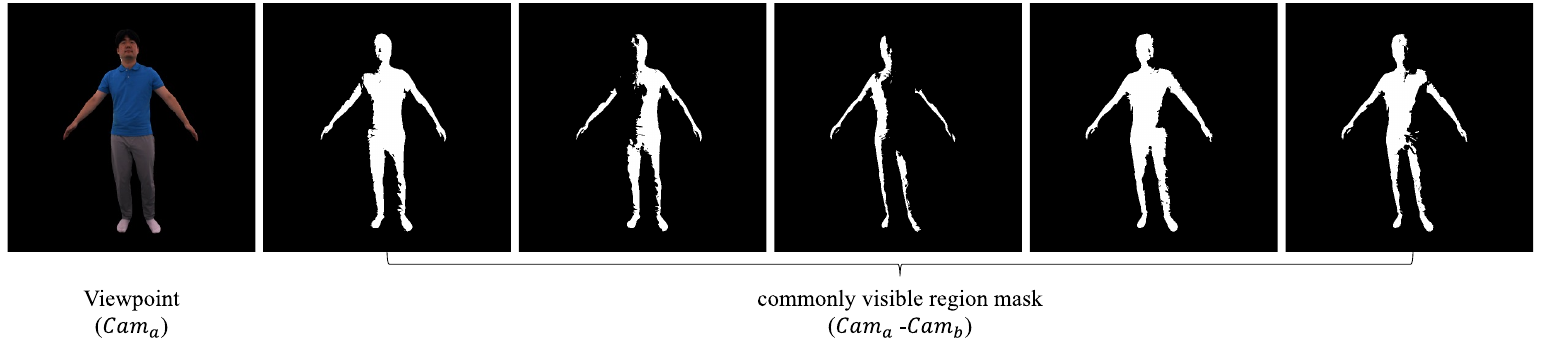}
	\caption{Example of visible region between $Cam_{a}$ and $Cam_{b}$.}
	\label{Fig:sub_mask}
\end{figure*}

\begin{figure*}[ht!]
	\centering
	\includegraphics*[width=\textwidth]{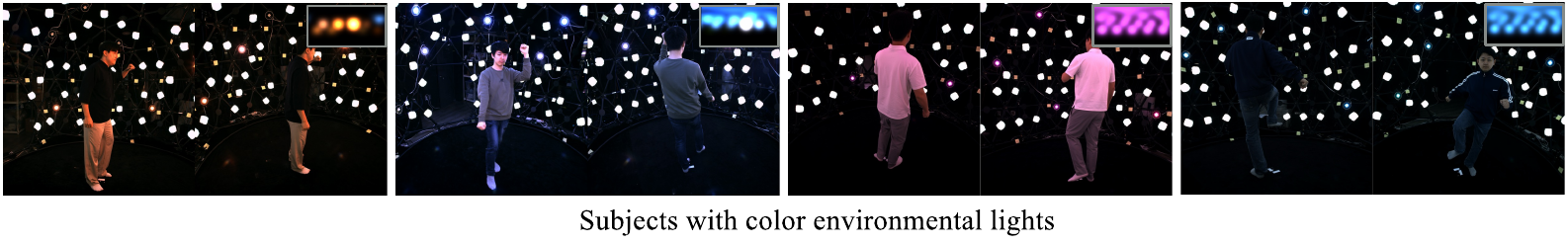}
	\caption{Examples of our constructed database.}
	\label{Fig:dataset}
\end{figure*}
\subsection{Preliminary}
\noindent \textbf{Animatable Avatar Modeling.} The human avatar modeling based on NeRF or 3DGS methods typically employ skeleton-driven deformation using a parametric human mesh (\eg SMPL or SMPL-X).
Given skinning weights $\mathcal{W}$ and joint transformation matrices $\{\theta_k\}^{J}_{k=1}$, where $J$ is the number of joints, the deformation is performed using the linear blend skinning (LBS) mechanism.
To do this, both approaches train the skinning weights for the deformation process.

Specifically, NeRF-based methods learn the skinning weights to transform points from the observation space to the canonical space (inverse skinning approach), whereas 3DGS-based methods regress skinning weights to transform points from the canonical space to the observation space (forward skinning approach).
As a result, a point $x_{c}$ in the cannonical space is transformed into a point $x_{o}$ the observation space as follows:
\begin{equation}
    \label{deformation}
        x_{o}= \mathcal{A}(\theta_{t}) \cdot x_{c} =\big(\sum^{J}_{k=1}\mathcal{W}_{k}\theta_{k}\big) \cdot x_{c},
\end{equation}
\noindent where, $\mathcal{A}(\theta)$ is a transformation matrices $\big[\mathcal{A}_{R}(\theta_{t});\mathcal{A}_{T}(\theta_{t})\big]$.
Additionally, in 3DGS-based methods, Gaussian attributes, such as the rotation $r_{c}$ in the canonical space, are transformed into the observation space as $r_{o}=\mathcal{A}_{R}(\theta_{t})r_{c}$.We note that the skinning weight in the deformation process is fixed weights.

\noindent \textbf{3D Gaussian Splatting.} 3D Gaussian Splatting (3DGS) explicitly represents 3D scenes by leveraging a set of 3D Gaussians, which are rendered through a rasterization process. To this end, a 3D Gaussian can be formulated as:
\begin{equation}
\label{gaussian}
    G(x) = e^{-\frac{1}{2}(x-\mu)^{T}\Sigma^{-1}(x-\mu)},
\end{equation}
\noindent where $\mu$ is mean, and $\Sigma$ represents 3D covariance matrix.
To ensure the positive semi-definiteness of $\Sigma$, $\Sigma$ is decomposed into quaternion vector $r\in \mathbb{R}^{4}$ and scale vector $s\in\mathbb{R}^{3}$. $r$ and $s$ are converted into a rotation matrix $R$, and a scaling matrix $S$, respectively. 
By using two matrices, $\Sigma$ is defined as $\Sigma=RSS^{T}R^{T}$.
The 3D Gaussians are projected onto the image plane through the splatting process to render the scene from a specific viewpoint. This requires a 2D covariance matrix in the image plane, which can be approximated using the 3D covariance matrix and the projection matrix. 

\begin{equation}
\label{sigma}
    \Sigma' = JW \Sigma W^{T}J^{T},
\end{equation}
\noindent where $W$ is a world-to-camera transformation matrix. $J$ represents an approximated projective transformation of Gaussian points. After projection, pixel colors are obtained through alpha blending. Specifically, we count the 2D Gaussians that overlap with each pixel and blend them as follows:
\begin{equation}
\label{alphablending}
    C = \sum_{i\in N}c_i\alpha_i \prod^{i-1}_{j=1} (1 - \alpha_i),
\end{equation}
\noindent where $c_i$, $\alpha_i$ are the color and density of $i$-th 2D Gaussian, respectively.

\begin{table*}[t]
    \centering
    \caption{Quantitative results of human avatar reconstruction on our database. ``$\uparrow$" indicates higher is better. ``$\downarrow$" indicates the opposition.}
    \resizebox{0.85\textwidth}{!}{
    \begin{tabular}{c|ccccc|ccccc|c}
    \hline\hline
    \multirow{2}{*}{Method} & \multicolumn{5}{c|}{Novel view} & \multicolumn{5}{c|}{Novel pose} & \multirow{2}{*}{TR} \\
                             & PSNR$\uparrow$   & SSIM$\uparrow$   & LPIPS$\downarrow$ & PSNR$_{n}$$\uparrow$   & SSIM$_{n}$$\uparrow$    & PSNR$\uparrow$   & SSIM$\uparrow$   & LPIPS$\downarrow$ & PSNR$_{n}$$\uparrow$   & SSIM$_{n}$$\uparrow$    \\ \hline
                            3DGS-Avatar\cite{qian20243dgs}                     &30.28&0.9683&0.0317&-&-      &30.12&0.9567&0.0352&-&-& 2h\\
                             GauHuman\cite{hu2024gauhuman}                      &31.34&0.9651&0.0305&-&-      &30.26&0.9516&0.0379&-&-& 1h\\
                             GoMAvatar\cite{wen2024gomavatar}                   &30.37&0.9689&0.0325&22.57&0.8035      &30.34&0.9688&0.0329&20.13&0.758& 12h\\
                             iHuman\cite{paudel2024ihuman}                      &28.21&0.9215&0.0342&19.81&0.7112      &24.84&0.9067&0.0402&17.65&0.6248& 2h\\
                             ExAvatar\cite{moon2024expressive}                  &31.42&0.9588&0.0171&-&-               &31.54&0.9414&0.0284&-&-& 2h\\
                             EVA\cite{hu2024expressive}                         &31.65&0.9514&0.0267&-&-               &30.15&0.9365&0.0317&-&-& 6h\\ \hline
                             Ours                                                &\textbf{33.85}&\textbf{0.9848}&\textbf{0.0115}&\textbf{29.58}&\textbf{0.9671}&\textbf{31.85}&\textbf{0.9748}&\textbf{0.0145}&\textbf{30.58}&\textbf{0.9606}& 6h\\ \hline \hline
    \end{tabular}
    }
    \label{table:quant_recon_zju}
\end{table*}

\subsection{Detail of Multi-view Multi-illuminated Dataset}
The key feature of our MvMi dataset is to capture human performance in high resolution under a wide range of colored lighting conditions. To do this, we established a system with  30 high-resolution cameras ($4096 \times 4096$). All cameras are positioned to capture a full $360^\circ$ view of a human subject. All cameras are synchronized using external triggers. For various lighting conditions, 19 custom-made LED modules are set up to simulate environmental lighting via Spherical Gaussian (SG) parameters. Each subject was recorded across eight distinct lighting scenarios, with varied poses in each sequence, resulting in approximately 11.5 million high-resolution multi-view video frames. Fig.~\ref{Fig:dataset} shows examples from our proposed dataset.

\subsection{Training procedure and Inference Pipeline}
We present inference pipeline of our method as shown in Fig.
We present the details of the training procedure and inference pipeline of our proposed method.
In the first stage, our method learns the articulation of avatars based on body motion by training the dynamic skinning weight, which is generated through the pose-dependent weight encoder. Here, $c_{s}$ is utilized for reconstruction guidance. In the second stage, we refine the appearance of avatars by employing PBR process. 

\subsection{Implementation details}
\noindent \textbf{Implementation details of Geometric Consistency Loss.}
To compute the geometric consistency loss, we need the intersection area on the avatar, which masks the common visible regions between two camera views.
In more detail, given two cameras $cam_{a}$ and $cam_{b}$, we can obtain the mask in the viewport of $cam_{a}$ through dot product between the normal and view direction between the avatar and $cam_{b}$.  The i-th Gaussian is invisible if $\hat{n}_i \cdot (x_{i} - cam_{b}) \leq t$.  We set $t=9^{\circ}$. We present an example of the masks in Fig~\ref{Fig:sub_mask}.

\noindent \textbf{Implementation details of training.}
Before setting the Gaussians on the vertices of SMPL~\cite{smpl}, we subsample the vertices to approximately 30K.
We do not perform Gaussian cloning, splitting, or pruning as in 3DGS~\cite{3dgs}.
We optimize the total objectives using the Adam optimizer with a learning rate of $1e^{-3}$. 
The batch size is set to 1, and training is conducted on a single NVIDIA A6000 Ada GPU for 6 hours.
We select one camera view from our dataset and use the first 4/5 of its frames as training data.
The remaining frames across all camera views are used to evaluate novel pose rendering.
For evaluating novel view rendering, we use the corresponding frames from all camera views except the selected view.

\section{Additional Results}
We conducted additional comparison on ZJU-MoCap database.
\noindent \textbf{Quantitative Results}
We show PSNR, SSIM, LPIPS, PSNR$_{n}$ and SSIM$_{n}$ on the ZJU-MoCap database in Tab~\ref{table:quant_recon_zju}.
As shown in the table, our method achieves high fidelity of both rendering quality and normal compared to the state-of-the-art methods.

\begin{figure*}[ht!]
	\centering
	\includegraphics*[width=0.8\textwidth]{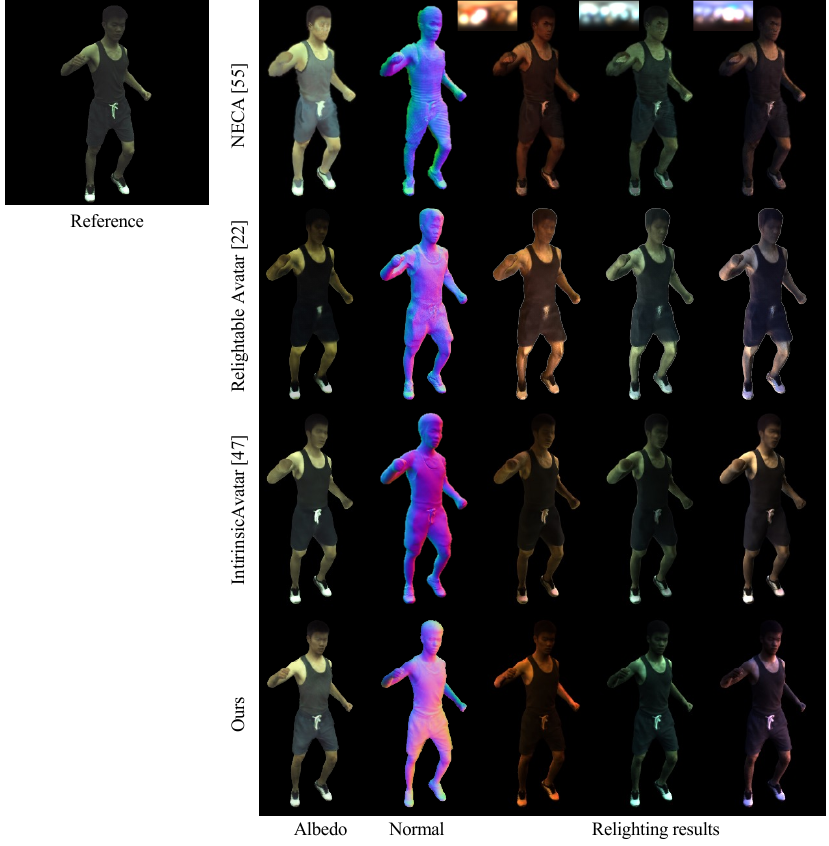}
	\caption{Qualitative results of human avatar relighting on ZJU-MoCap database.}
	\label{Fig:qa_relight1}
\end{figure*}

\noindent \textbf{Qualitative Results}
We present a qualitative result on the ZJU-MoCap dataset to demonstrate the reconstruction as shown in Fig.~\ref{Fig:qa_recon1}. Our proposed framework can produce photo-realistsic human avatar with fine-grained geoemtry. This ensure the photo-realistsic rendering results under arbitrary lighting environment as shwon in Fig.~\ref{Fig:qa_relight1}.

\begin{figure*}[ht!]
	\centering
	\includegraphics*[width=0.7\textwidth]{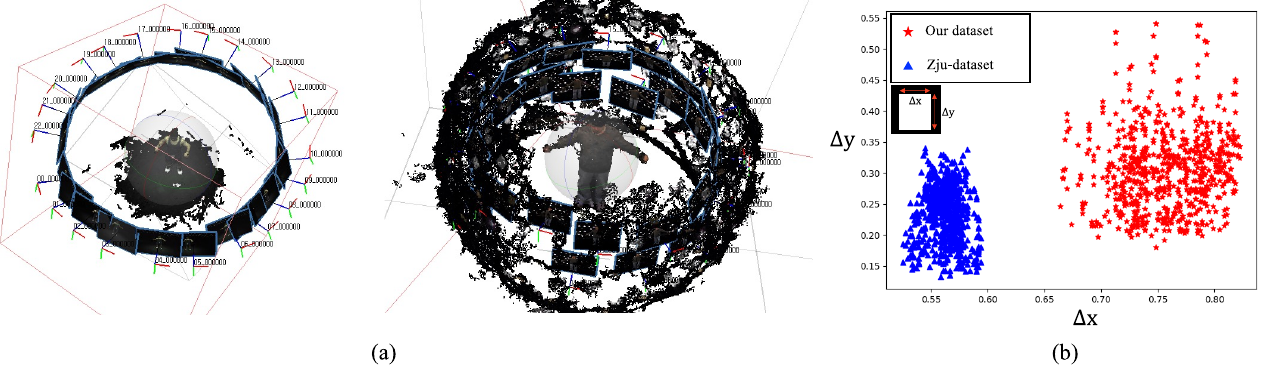}
	\caption{Comparison of (a) the capture systems and (b) the bounding box width and height distributions between the ZJU Mocap~\cite{peng2021neural} and our proposed dataset.}
	\label{Fig:db}
\end{figure*}

\noindent \textbf{Additional Analysis}
As shown in Tables~\ref{table:quant_recon} and~\ref{table:quant_recon_zju}, we observe that the quantitative performance is slightly lower when trained on our proposed dataset compared to the ZJU dataset. To investigate this discrepancy, we conducted a detailed analysis of the differences between the two datasets.
Fig.~\ref{Fig:db} (a) illustrates the differences in capture environments: the left side shows the camera setup used in the ZJU dataset, while the right side depicts the camera arrangement in our proposed dataset. Notably, the subject-to-camera distance in the ZJU dataset is shorter than in our setup, suggesting that the range of motion appears more compact in the captured images. Additionally, we explicitly analyzed the differences between the two datasets. To ensure a fair comparison, we resized all images from both datasets to a resolution of $512 \times 512$ and extracted the subject’s bounding box in each frame. We then measured the width ($\Delta x$) and height ($\Delta y$) of each bounding box and normalized these values by dividing them by 512.
The results are shown in Fig.~\ref{Fig:db} (b). We observe that, in general, the bounding box size in our dataset is larger compared to that in the ZJU dataset, leading  to a wide range of variations in the image domain. This difference may affect the visual motion cue, contributing to the performance gap.

\section{Discussion}
Our proposed framework can reconstruct a high-fidelity human avatar and also enables rendering the avatar under arbitrary lighting conditions using PBR process. However, PBR process is difficult to present complex reflectance. While ray tracing could address this, its high computational complexity led us to use a rough approximation. To mitigate these limitations, we considers the two most important dimensions: (1) a well-constructed dataset, and (2) a well-designed modeling framework. Still, our database can benefit more from more diverse reflectance scenarios, and our framework can be equipped with better generative models such as diffusion models.


\end{document}